\documentclass[10pt,twocolumn,letterpaper]{article}

\usepackage[pagenumbers]{cvpr} 

\definecolor{cvprblue}{rgb}{0.21,0.49,0.74}
\usepackage[pagebackref,breaklinks,colorlinks,allcolors=cvprblue]{hyperref}

\usepackage{csquotes}
\usepackage{mathtools}
\usepackage{tabularx}
\usepackage{array}
\usepackage{algorithm}
\usepackage{algpseudocode}
\usepackage{multirow}
\usepackage{makecell}
\usepackage{listings}

\def\paperID{7749} 
\def\confName{CVPR}
\def\confYear{2025}

\title{Interpretable Image Classification via Non-parametric Part Prototype Learning}

\newcommand*{\affmark}[1][*]{\textsuperscript{#1}}

\author{Zhijie Zhu\affmark[1]\thanks{Equal contribution.} \quad Lei Fan\affmark[1]\footnotemark[1] \quad Maurice Pagnucco\affmark[1] \quad Yang Song\affmark[1]\\
\affmark[1]University of New South Wales\\
{\tt\small \{zhijie.zhu@,lei.fan1@,morri@cse,yang.song1@\}.unsw.edu.au}
}

\begin{document}

\maketitle
\begin{abstract}
  Classifying images with an interpretable decision-making process is a long-standing problem in computer vision. In recent years, Prototypical Part Networks has gained traction as an approach for self-explainable neural networks, due to their ability to mimic human visual reasoning by providing explanations based on prototypical object parts. However, the quality of the explanations generated by these methods leaves room for improvement, as the prototypes usually focus on repetitive and redundant concepts. Leveraging recent advances in prototype learning, we present a framework for part-based interpretable image classification that learns a set of semantically distinctive object parts for each class, and provides diverse and comprehensive explanations. The core of our method is to learn the part-prototypes in a non-parametric fashion, through clustering deep features extracted from foundation vision models that encode robust semantic information. To quantitatively evaluate the quality of explanations provided by ProtoPNets, we introduce Distinctiveness Score and Comprehensiveness Score. Through evaluation on CUB-200-2011, Stanford Cars and Stanford Dogs datasets, we show that our framework compares favourably against existing ProtoPNets while achieving better interpretability. Code is available at: \url{https://github.com/zijizhu/proto-non-param}.
\end{abstract}
    
\section{Introduction}
\label{sec:intro}

\begin{figure}
  \centering
  \vspace{-1em}
  \includegraphics[width=0.5\textwidth]{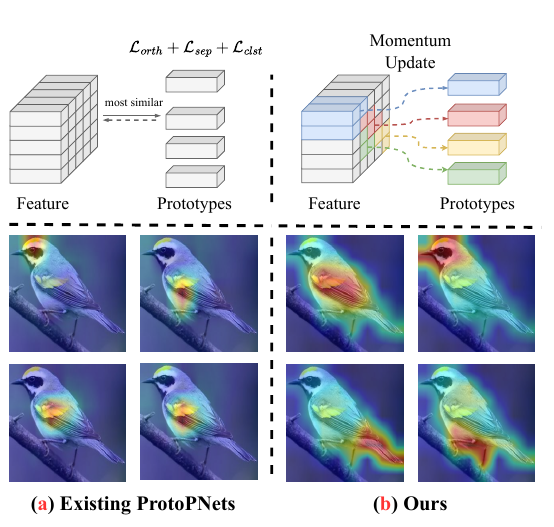}
  \vspace{-1em}
  \caption[architecture]{\textbf{a.} Existing methods \cite{wang_interpretable_2021,huang_evaluation_2023} encode various assumptions as regularizations to guide prototype learning, but often fail to diversify the concepts learned by each prototype. Here multiple part-prototypes attend to the same image region, thereby limiting their interpretability. In contrast \textbf{b.} Our method partitions the feature space into semantically distinct clusters, and updates each prototype with the empirical mean of its respective cluster, enabling each prototype to learn semantically different concepts.}
  \label{fig:comparison}
  \vspace{-1em}
\end{figure}

In recent years, self-explainable networks have emerged as a promising approach for improving the transparency of deep neural networks by providing inherent interpretability. Unlike traditional explanation methods that operate post-hoc \cite{selvaraju_grad-cam_2017,zhou_learning_2016}, recent approaches aim to mimic human reasoning by justifying predictions through an interpretable decision-making process. Examples include Prototypical-Part Networks (ProtoPNets) \cite{chen_this_2019,donnelly_deformable_2022}, part discovery methods \cite{van_der_klis_pdisconet_2023,huang_interpretable_2020}, and Concept Bottleneck Models \cite{koh_concept_2020,wang_learning_2023,espinosa_zarlenga_concept_2022, zhang2024decoupling}, which offer explanations derived from object concepts. Among these, ProtoPNets have garnered attention as they provide intuitive visual explanations rooted in past examples. Specifically, they define a number of prototypes for each object category, each representing a distinguishable object part, and make the prediction based on the similarity of an input image to prototypes, thereby emulating the process of human visual reasoning.

However, existing ProtoPNets train prototypes as learnable parameters, and guide their learning process with the core assumption that a region within each input image should match one of the prototypes from the corresponding category. However, we observe that multiple prototypes from the same class often correspond to the same object part, which hinders the quality of explanations provided by the network in two ways: First, explanations become non-comprehensive, covering only a subset of object parts. Second, this results in redundant and repetitive explanations that fail to adequately illustrate why the object is classified in a particular way. An example is shown in Figure \ref{fig:comparison}.a, out of the four most activated prototypes in a ProtoPNet, three of them focus on overlapping regions near the bird's wing, ignoring other parts such as the bird's legs and tail. Ideally, within an interpretability framework, the model is expected to recognize and differentiate a comprehensive set of distinctive parts for each object category, thus enhancing the clarity and transparency of its decision-making process.

To approach this problem, we take inspiration from recent advances in prototype-based classification \cite{zhou_prototype-based_2024,yang_prototype_2020}, and encode distinct object parts using \textbf{non-parametric} prototypes \cite{zhou_rethinking_2022,wang_visual_2023}, as shown in Figure \ref{fig:comparison}.b. Non-parametric prototypes were originally proposed to learn a more robust feature space by capturing the rich variance within each class through multiple prototypes \cite{zhou_rethinking_2022}. We extend this idea to part-based interpretability to address both of the aforementioned limitations. Specifically, in comparison to previous methods that guide the learning process of learnable prototypes with regularizations \cite{wang_interpretable_2021}, non-parametric prototypes allow finer control over the representation of each prototype within the same class by applying constraints on feature clustering. As a consequence, each prototype is able to capture unique characteristics of an object that differentiate it from others, resulting in \textbf{semantically distinct} prototypes, each latent feature patch of the foreground object is able to contribute to one of the prototypes, allowing the overall representation for each class to embody a \textbf{comprehensive} set of associated concepts. These properties greatly improve the interpretability of the explanations.

Additionally, in order to obtain robust feature representations that encode object part semantics, We explore advanced foundation models as backbones to extract features, specifically utilizing self-supervised Vision Transformers pre-trained with self-distillation (DINO-ViTs) \cite{caron_emerging_2021,oquab_dinov2_2024}, given that these backbone networks capture rich semantic information about object parts and can be used out-of-the-box \cite{park_what_2022,amir_deep_2022}. We also propose an efficient fine-tuning strategy for self-supervised ViT backbones to adapt the feature space to the target domain while grounding the embedding of object parts to its assigned prototypes. Hence, our framework consists of two training stages where the initial stage focuses on prototype discovery, and the subsequent stage is directed towards feature refinement. Only the image-level class labels are required in both stages. We conduct experiments to compare our approach with existing ProtoPNets on the CUB-200-2011 \cite{WahCUB_200_2011}, Stanford Cars \cite{krause_3d_2013}, and Stanford Dogs datasets \cite{dataset2011novel}. Moreover, to evaluate the diversity of concepts captured by ProtoPNets, we introduce two new metrics:  \textit{Distinctiveness Score}, which quantifies the extent of overlap among object parts learned by prototypes within the same class, and \textit{Comprehensiveness Score}, which measures whether the prototypes comprehensively attend to different parts of the foreground object. To summarize the key contributions of this work:
\begin{itemize}
  \item We identify a key limitation of existing ProtoPNets, specifically generating repetitive and redundant visual explanations. To this end, we leverage non-parametric prototypes to build a self-explainable network.
  \item Our method comprehensively explores non-parametric prototypes and foundation vision backbones, discovering diverse and semantic distinctive concepts.
  \item We introduce two evaluation metrics, namely Distinctiveness Score and Comprehensiveness Score to quantify the diversity and comprehensiveness of object part concepts. We demonstrate that our method yields comparable results against state-of-the-art ProtoPNets models on three benchmarking datasets while recognizing holistic and distinctive concepts from each category.
\end{itemize}


\section{Related Work}
\label{sec:related_work}

\noindent \textbf{Part-based Interpretable Image Classification.} Drawing inspiration from the human ability to distinguish objects by recognizing their semantic parts, the Prototypical Part Network (ProtoPNet) \cite{chen_this_2019} emulates this reasoning process by comparing image patches with learned prototype vectors that represent object parts specific to each class. Several variants have been proposed to enhance both its interpretability and classification accuracy \cite{donnelly_deformable_2022, tan_post-hoc_2024, carmichael_pixel-grounded_2024, xue_protopformer_2022}. Notably, TesNet \cite{wang_interpretable_2021} reformulates prototype learning from the perspective of constructing embedding space on Grassmann Manifold, and also improves the overall performance by applying various priors on prototypes, such as orthonormality. Building upon the architecture of TesNet, Huang \textit{et. al} \cite{huang_evaluation_2023} introduced modules to improve consistency and stability of the visual explanations, as well as corresponding metrics for quantitative evaluation.

Meanwhile, there has been a parallel line of research that focuses on detecting object parts of fine-grained object categories \cite{van_der_klis_pdisconet_2023,huang_interpretable_2020}. While these methods involve learning object part representations, they usually detect a number of shared object parts for the entire target domain and emphasize evaluating the quality of detected parts rather than leveraging them for case-based reasoning.

The main goal of our work aims to alleviate aforementioned drawbacks of ProtoPNets, by learning part-prototypes in a non-parametric fashion. In addition, Huang et. al. \cite{huang_concept_2024} integrate the reasoning process of ProtoPNets with the series of Concept Bottleneck Models \cite{koh_concept_2020,espinosa_zarlenga_concept_2022} to improve the trustworthiness of explanations \cite{huang_concept_2024}. We follow this pioneering work and evaluate our framework when adapted to the concept learning setting.

\noindent \textbf{Foundation Vision Models.} Recently, foundation vision backbones \cite{caron_emerging_2021,radford_learning_2021} have been successfully applied to various challenging downstream tasks \cite{fan2022cancer,awais_foundational_2023,fan2025grainbrain,kirillov_segment_2023}, yielding remarkable results. In contrast to traditional supervised training, these backbones typically employ self-supervision or weak supervision schemes, such as masked image modeling \cite{xie_simmim_2022}, contrastive learning \cite{chen_simple_2020} or self distillation, reducing the reliance on labels and enhancing the quality of representation. Previous studies have demonstrated that each of these pre-training methods constructs an embedding space with different desirable properties \cite{park_what_2022, xie_revealing_2023}. In particular, the attention map and patch tokens of self-supervised Vision Transformers trained with self-distillation have been shown to encode rich semantic information, such as scene layouts, which are beneficial for dense recognition tasks like and object discovery \cite{simeoni_localizing_2021}, anomaly detection \cite{fan2023av4gainsp,fan2023identifying}, and image generation \cite{wang2024towards,sun2024eggen}. In addition, other works have leveraged their strong object part representation for fine-grained categorization \cite{saha_particle_2023,amir_deep_2022,fan2022grainspace}. Building upon these developments, our framework harnesses these desired properties to improve the quality of explanation provided by part-based interpretable image classifiers.

\noindent \textbf{Prototype Learning.} Prototype-based classification is to make predictions based on representative examples from past data, similar to how humans approach problem solving and learning, as suggested by psychology studies \cite{aamodt_case-based_1994}. Recently, it has gained significant interest in deep learning as an alternative to traditional classification heads. For example, memory bank based methods \cite{he_momentum_2020,wu_unsupervised_2018} retain a large number of past samples to learn high-quality representations. Approaches such as ProtoSeg \cite{zhou_rethinking_2022}, deep nearest centroids \cite{wang_visual_2023} and various segmentation models \cite{lu_promotion_2024,qin_unified_2023} replace conventional parametric softmax layers of segmentors with multiple non-parametric prototypes, which not only learn class distributions but also capture the intra-class variance of each category, improving the representativeness and generalizability of the models. Our non-parametric part-prototype is inspired by these approaches.



\section{Method}
\label{sec:method}

\begin{figure*}[t]
  \centering
  \includegraphics[width=1.0\textwidth]{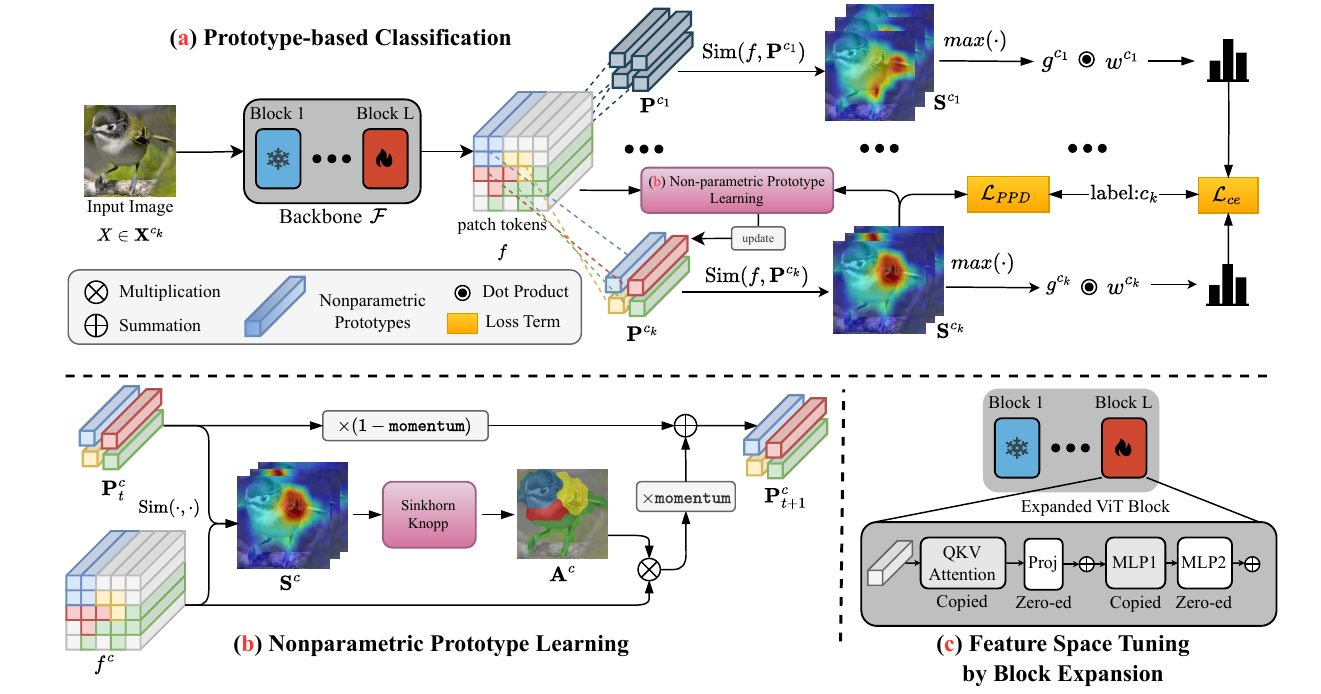}
  \caption[architecture]{The architecture of our learning framework. \textbf{a.} Similar to prototype-based classification, we define $K$ number of part-prototypes for each category and compare them to the latent features, which yields similarity maps. Each similarity map is pooled to generate a prototype presence vector for each class. The final logits are computed as a weighted average of the presence vector. \textbf{b.} Non-parametric Prototype Update: We employ feature clustering to discover class-wise non-parametric part prototype from latent feature patches. Each part prototype can be regarded as the empirical mean of its corresponding part from all training examples. \textbf{c.} The feature space of the backbone is fine-tuned efficiently with part-prototype fixed, and by inserting new ViT blocks.}
  \label{fig:network-arch}
  \vspace{-1em}
\end{figure*}

Part-based interpretability methods such as ProtoPNets aim to explain the predictions by generating final logits based on how much input image regions look like prototypical object parts from the training set. The core of these methods is to learn a robust set of prototypes that can provide high quality explanations. As shown in Figure \ref{fig:network-arch}.a, our method achieves this by deriving them as non-parametric prototypes with a learning procedure disentangled from the classification process. This results in two training stages, where the first stage is dedicated to train the prototypes and the second stage focuses on refining the feature space.

\subsection{Overview}

The goal of our model is to perform interpretable image classification by learning prototypes $\mathbf{P} \in \mathbb{R}^{C \times K \times D}$, where $C$ is the number of classes and $K$ is the number of prototypes belonging to each class. Each $D$ dimensional vector represents one of the $K$ representative parts belonging to class $c \in \{c_1, c_2, ...\}$. During inference, given a batch of image $\mathbf{X}_{1:B} = \{\mathbf{X}_b \in R^{H \times W \times 3} \}$, where $B$ is the batch size, $H$ and $W$ denote the size of the input image, a ViT backbone $\mathcal{F}$ first encodes each image into a sequence of patch tokens $\mathbf{f} = \mathcal{F}(\mathbf{X}) \in \mathbb{R}^{B \times h \times w \times D}$, reshaped into a feature map of shape $h \times w$. Our model computes the similarity between the patch tokens $\mathbf{f}$ and part prototypes $\mathbf{P}$, generating a set of similarity maps $\mathbf{S}_{1:B} = \{\text{Sim}(\mathbf{f}_b, \mathbf{P}) \in \mathbb{R}^{h \times w \times C \times K}\}_{b=1}^{B}$, where $\text{Sim}(\cdot, \cdot)$ is a similarity measure, such as cosine similarity. As shown in Figure \ref{fig:network-arch}.a, each similarity map highlights a region of the input image similar to the learned prototypical part. Finally, the maximum value of each similarity map quantifies the occurrence of its corresponding part prototype with respect to the input, and the logit of class $c$ is computed as the weighted average of all occurrence scores of its prototypes.

In the first stage of training, an efficient weakly-supervised clustering algorithm computes non-parametric part prototypes for each class based on extracted features from frozen backbones. Due to the well-structured latent space, these prototypes can be directly used for interpretable image classification. Then, we fix the learned part prototypes and fine-tune the backbone network by using the prototypes as anchors. This adapts the feature space to the target domain and significantly enhances the classification performance. We only use image-level class labels for supervision in both stages.

\subsection{Non-parametric Prototype Learning}
As shown in Figure \ref{fig:network-arch}.b, the main idea of this stage is to cluster the feature representation of each class involved in a batch of samples, and create a mapping between each patch token and part prototype. The empirical mean of the patch tokens assigned to the same part-prototype is directly used to update the corresponding prototypes. Hence, each latent feature patch is able to contribute to one of the part-prototypes, ensuring the overall representation of each class to cover a wide range of concepts. The only supervision involved in this stage is associating each patch token with the class label of image it belongs to, and the assignment maps are computed in an unsupervised manner.

Specifically, we treat a batch of latent patch tokens as a bag of features, reshaping the patch tokens $\mathbf{f}$ and similarity maps $\mathbf{S}$ to $\mathbf{f} \in \mathbb{R}^{Bhw \times D}$ and $\mathbf{S} \in \mathbb{R}^{Bhw \times C \times K}$. Next, we employ a foreground extraction module to extract patches belonging to the foreground of each sample, excluding the background patches to improve representation quality. Given $N$ foreground patches of class $c$, denoted as $\mathbf{f}^{c} \subset \mathbf{f}, \quad \mathbf{f}^{c} \in \mathbb{R}^{N \times D}$, we obtain a similarity matrix $\mathbf{S}^{c} \in \mathbb{R}^{N \times K}$ between $\mathbf{f}^{c}$ and $\mathbf{P}^{c}$ by indexing $\mathbf{S}$. We then compute a one-hot cluster assignment $\mathbf{A} \in \{0, 1\}^{N \times K}$ that assigns each patch of class $c$ to one of the part-prototypes.

In particular, we aim to optimize $\mathbf{A}^c$ so that the similarity between each patch and its assigned prototype is maximized, that is:

\begin{equation}
  \label{eq:1}
  \max_{\mathbf{A}^{c} \in \mathcal{A}^c} \texttt{Tr}((\mathbf{A}^c)^{\top} \mathbf{S}^c) ,
\end{equation}
where $\texttt{Tr}(\cdot)$ is the trace of a matrix. To prevent trivial solutions for $\mathbf{A}^c$ and ensure each cluster represents a distinct object part, we enforce \textit{unique assignment} and \textit{equipartition constraints} \cite{zhou_rethinking_2022}:

\begin{equation}
  \label{eq:2}
  {(\mathbf{A}^c)}^{\top}\mathbf{1}_{N}=\frac{1}{K}\mathbf{1}_{K}, \mathbf{A}^{c} \mathbf{1}_{K}=\frac{1}{N}\mathbf{1}_N .
\end{equation}

This ensures each patch token maps to a single part-prototype of class $c$, and each prototype has an approximately equal number of patch tokens assigned. This formulation can be regarded as an Optimal Transport (OT) problem, where the approximate solution of $\mathbf{A}$ can be estimated by solving a smoothed version of equation \eqref{eq:1}:

\begin{equation}
  \label{eq:3}
  \mathbf{A}^{c,*} = \max_{\mathbf{A}^c \in \mathcal{A}^c} \texttt{Tr}((\mathbf{A}^c)^{\top} \mathbf{S}) + \kappa h(\mathbf{A}^c) ,
\end{equation}
where $\kappa h(\mathbf{A})$ is the entropic regularization term controlling smoothness. We use $\kappa=0.05$ following previous works \cite{cuturi_sinkhorn_2013}. The solution can be efficiently computed using the Sinkhorn-Knopp Algorithm:

\begin{equation}
  \label{eq:4}
  \mathbf{A}^{c,*} = \texttt{diag}(\mathbf{u}) \text{exp}\left(\frac{\mathbf{S}^c}{\kappa}\right) \texttt{diag}(\mathbf{v}).
\end{equation}

\noindent \textbf{Momentum Update of Prototypes.} Given the estimated one-hot patch-prototype assignment $\mathbf{A}^{c,*}$, each part-prototype $\mathbf{P}^{c}_{k}$ is updated as the mean of its assigned patches:
\begin{equation}
  \label{eq:5}
  \mathbf{P}^{c}_{k} \leftarrow \mathbf{P}^{c}_{k} + (1 - \beta)\overline{\mathbf{f}}^{c,k} ,
\end{equation}
where $\beta \in [0,1]$ is a coefficient that controls the update speed, and $\overline{\mathbf{f}}$ is the mean feature. This learning scheme effectively captures the mean embedding of object parts across all training examples, and can be visualized by projecting to the closest patches from the training set \cite{chen_this_2019}.

\subsection{Adapt to Interpretable Image Classification}

As shown in Figure \ref{fig:network-arch}.a, the learned part-prototypes can be directly used for classification. The activation of each prototype $\mathbf{P}^{c}_{k}$ with respect to the input image is calculated as:

\begin{equation}
  \label{eq:6}
  g^{c}_{k} = \max_{\widetilde{\mathbf{f}} \in \mathbf{f}} \quad \text{Sim}(\widetilde{\mathbf{f}}, \mathbf{P}^{c}_k) \in \mathbb{R}.
\end{equation}

As not all prototypes are equally important for classifying the input image, following previous work \cite{huang_evaluation_2023}, we generate the final logit for each class $c$ by computing a weighted average of $g^{c}$:

\begin{equation}
  \label{eq:7}
  \text{logit}^{c} = \sum^{K}_{k=1} w^{c}_{k} \cdot g^{c}_{k},
\end{equation}

\noindent where $w$ is a learned modulation vector that controls the contribution of each prototype activation towards the logit of its corresponding class. In contrast to using a fully-connected layer, computing class logits as a weighted average ensures that prototypes contribute exclusively to their respective class, thereby enhancing the interpretability.

\noindent \textbf{Prototype-Anchored Feature Space Tuning.}
After discovering class-wise part-prototypes from the pre-trained backbone, we fine-tune the backbone to ground the features to the learned prototypes and adapt to the target domain. To this end, we employ Block Expansion \cite{bafghi_parameter_2024}, as shown in Figure \ref{fig:network-arch}.c. Given a ViT backbone $\mathcal{F}$ with a sequence of transformer blocks $(\mathcal{F}_{1}, \mathcal{F}_{2}, \mathcal{F}_{3}, ...)$, Block Expansion divides the blocks into $m$ groups and inserts a new identity block $\mathcal{F}_{id}$ at the end of each divided group. Each inserted identity block is initialized with weights of the preceding block, with the weights of attention projection layer and the second MLP layer set to zero, such that $\mathcal{F}_{id}(x) = x$, \textit{i.e.}, the output of the backbone remains unchanged without further training. During fine-tuning, we freeze all original weights and part-prototypes, and only fine-tune the added identity blocks and modulation vectors. This approach efficiently produces features that separate classes well without disturbing the well-structured latent space of the backbone.

\noindent \textbf{Patch-Prototype Distance Optimization.}
In the fine-tuning stage, we use a standard cross-entropy loss $\mathcal{L}_{ce}$ to separate features of each class. In addition, we ground each image patch to its assigned part-prototype to improve feature locality. To this end, we introduce an in-class patch-prototype contrastive loss, defined as:

\begin{equation}
  \label{eq:8}
  \mathcal{L}_{\text{PPC}} = \sum_{c \in \mathcal{C}} \frac{\text{exp}((\mathbf{f}^{c}_{i})^{\top}\mathbf{P}^{c}_{k_i})}{\text{exp}((\mathbf{f}^{c}_{i})^{\top}\mathbf{P}^{c}_{k_i})+\sum_{\mathbf{P}^{c}_{k_j} \in \mathcal{P}^{c-}}\text{exp}((\mathbf{f}^{c}_{i})^{\top} \mathbf{P}^{c}_{k_j})},
\end{equation}
where $\mathcal{P}^{c-} = \{\mathbf{P}^{c}_{k}\}^{K}_{k=1} / \mathbf{P}^{c}_{k_i}$. This loss \enquote{pulls} each patch towards its assigned part-prototype in the latent space, and separate it from other part-prototype within the same class. Hence, we jointly train the expanded ViT block and the modulation weights with the following objective:

\begin{equation}
  \label{eq:9}
  \mathcal{L}_{\text{total}} = \mathcal{L}_{\text{ce}} + \lambda_{\text{PPC}} \mathcal{L}_{\text{PPC}}.
\end{equation}

\subsection{Distinctiveness Score}

Ideally, each part-prototype belonging to the same class should learn a distinct concept from objects; otherwise, the explanation would be confusing and redundant. To quantify this property, we introduce a metric, Distinctiveness Score that measures the amount of overlap between image regions highlighted by the part prototypes associated with the ground truth class of the input image.

Specifically, given an input image $x$ of class $c_x$ and the corresponding activation maps $S^{c_x} \in \mathbb{R}^{h \times w \times K}$ generated by prototypes $\mathbf{P}^{c_x}$, we first resize the activation maps to $\hat{S}^{c_x} \in \mathbb{R}^{K \times H \times W}$ to match the size of input $x$. Next, we extract the object region $R^{c_x,k}(x)$ highlighted by each activation map $S^{c_x,k}$ as a fix-sized bounding box with predetermined height ${H_{b}}$ and width $W_{b}$ around its max activation. The amount of overlap $O(x)$ between the similarity maps $S^{c_x}$ generated by ground truth class prototypes is then quantified by the Intersection over Union (IoU) between each pair of highlighted regions $R^{c_x,k}(x)$, as follows:
\begin{equation}
  \label{eq:10}
  O(x) = \frac{1}{K(K-1)/2}\sum_{k=1}^{K-1} \sum_{l=k+1}^{K} \text{IoU}(R^{c_x,k}(x), R^{c_x,l}(x)).
\end{equation}

The Distinctiveness Score is computed based on the average overlap across all samples in the test set $\mathcal{X}$, and a higher value is better. It is defined as:
\begin{equation}
  \label{eq:11}
  1 - \frac{1}{|\mathcal{X}|} \sum_{x \in \mathcal{X}} O(x) .
\end{equation}

\subsection{Comprehensiveness Score}

\begin{figure}[t]
  \centering
  \includegraphics[width=1.0\columnwidth]{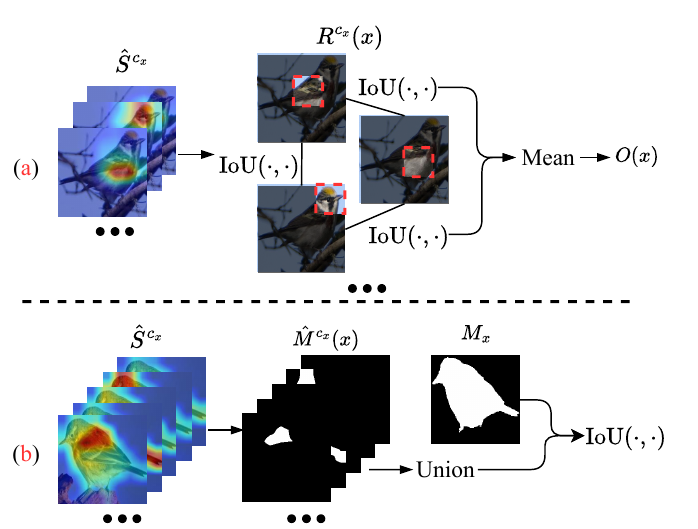}
  \caption{\textbf{a.} The \textbf{Distinctiveness} score for one sample is calculated based on the amount of overlap between areas attended by prototypes. \textbf{b.} The \textbf{Comprehensiveness} score is computed by comparing the ground truth foreground mask $M_x$ with the union of thresholded image region attended by prototypes.}
  \label{fig:score-calculation}

\end{figure}

A good Distinctiveness Score by itself does not guarantee good interpretability, as a high score could be achieved by networks with prototypes correspond to different image regions that lack meaningful concepts, e.g., background patches that contribute less to object recognition. Therefore, we further introduce Comprehensiveness Score to evaluate how thoroughly the object foreground is attended by the prototypes. Given an input image $x$ and the corresponding activation maps $S^{c_x}$ generated by prototypes $\mathbf{P}^{c_x}$, we perform bin-max normalization on each $S^{c_x,k}$ and obtain a mask $\hat{M}^{c_x,k}_i(x)$ of the most activated region of $\mathbf{P}^{c_x,k}$ using a threshold $\tau$:
\begin{equation}
  \label{eq:12}
  \tilde{S}^{c_x,k}_i = \frac{S^{c_x,k}_i - S^{c_x,k}_{\min}}{S^{c_x,k}_{\max} - S^{c_x,k}_{\min}} \text{,} \quad
  \hat{M}^{c_x,k}_i(x) =
  \begin{cases}
    1 & \text{if } \tilde{S}^{c_x,k}_i \geq \tau \\
    0 & \text{otherwise}.
  \end{cases}
\end{equation}

We then compute the IoU between the union of all masks $\hat{M}^{c_x,k}(x)$ and the ground truth object foreground mask $M_x$. The Comprehensiveness Score is calculated as the average IoU across all test samples $\mathcal{X}$, and a higher value indicates better comprehensiveness. Specifically, it is defined as:
\begin{equation}
  \label{eq:13}
  \frac{1}{|\mathcal{X}|} \sum_{x \in \mathcal{X}} \text{IoU}(M_x,\bigcup_{k=1}^K \hat{M}^{c_x,k}(x)).
\end{equation}

\begin{table*}[t]
  \centering
  \resizebox{\textwidth}{!}{
    \begin{tabular}{l||ccc|c||ccc|c||ccc|c}
      & \multicolumn{4}{c||}{\textbf{DINOv2 ViT-B/14}} & \multicolumn{4}{c||}{\textbf{DINOv2 ViT-S/14}} & \multicolumn{4}{c}{\textbf{DINO ViT-B/16}} \\
      & Con. & Sta. & Dis. & Class.
                    & Con. & Sta. & Dis. & Class.
                                  & Con. & Sta. & Dis. & Class.\\
      \hline
      \hline
      ProtoPNet (K=3) & 0.77 & 68.40 & 5.77 & 88.47 & 4.84 & 74.68 & 8.83 & 84.90 & 3.00 & 63.87 & 3.14 & 75.94 \\
      ProtoPNet (K=5) & 1.30 & 63.84 & 11.99 & 88.16 & 4.12 & 70.10 & 16.73 & 85.52 & 3.02 & 46.34 & 6.79 & 78.91 \\
      \hline
      Deformable ProtoPNet (K=3) & 0.95 & 80.00 & 5.90 & 86.99 & 9.67 & 61.18 & 1.65 & 83.09 & 1.55 & 70.84 & 9.15 & 71.49 \\
      Deformable ProtoPNet (K=5) & 2.83 & 80.43 & 10.34 & 88.14 & 11.73 & 79.87 & 4.76 & 85.54 & 0.95 & 73.16 & 1.17 & 81.74 \\
      \hline
      ProtoPool (K=3) & 29.83 & 52.83 & 87.59 & 88.72 & 15.50 & 47.89 & 91.17 & 86.19 & 26.67 & \textbf{77.74} & 83.89 & 72.04 \\
      ProtoPool (K=5) & 29.27 & 57.49 & 89.62 & 89.90 & 12.1 & 52.44 & 92.26 & 87.41 & 37.40 & 52.73 & 87.66 & 78.18 \\
      \hline
      TesNet (K=3) & 36.33 & 66.45 & 41.24 & 90.66 & 58.00 & 66.89 & 27.80 & 88.23 & 35.50 & 65.40 & 58.85 & 81.24 \\
      TesNet (K=5) & 58.60 & 74.60 & 81.93 & 90.35 & \textbf{67.80} & 70.27 & 81.16 & 88.95 & \textbf{51.30} & 70.81 & 81.94 & 82.29 \\
      \hline
      EvalProtoPNet (K=3) & \textbf{67.83} & 71.39 & 25.67 & 89.73 & 45.17 & 63.51 & 17.01 & 87.66 & 37.83 & 69.38 & 35.10 & 84.21 \\
      EvalProtoPNet (K=5) & 67.80 & 74.40 & 72.12 & 90.06 & 54.60 & 70.27 & 77.41 & 88.19 & 51.09 & 66.86 & 72.74 & 84.54 \\
      \hline
      Ours (K=3) & 61.17 & 78.98 & 86.44 & 90.37 & 61.50 & 75.57 & 89.13 & \textbf{89.07} & 45.67 & 69.90 & 95.62 & 85.12 \\
      Ours (K=5) & 66.40 & \textbf{82.97} & \textbf{94.45} & \textbf{90.82} & 66.01 & \textbf{81.31} & \textbf{95.75} & 88.16 & 47.20 & 75.57 & \textbf{96.59} & \textbf{86.93} \\
    \end{tabular}
  }
  \caption{Consistency score (\%), Stability Score (\%), Distinctiveness Score (\%) and classification accuracy (\%) of various ProtoPNet on CUB-200-2011. K denotes the number of prototypes defined for each class. We train previous methods with their official implementations.}
  \label{tab:main-protopnets-results}
\end{table*}




\section{Experiments}
\label{sec:experiments}

\subsection{Experimental Settings}

\noindent \textbf{Datasets.} To evaluate the classification accuracy and interpretability of our proposed framework, we conduct experiments on CUB-200-2011 \cite{WahCUB_200_2011}, Stanford Cars \cite{krause_3d_2013} and Stanford Dogs \cite{dataset2011novel} dataset, following previous works on ProtoPNets. The CUB-200-2011 dataset consists of 11,788 images spanning 200 bird species. Stanford Cars includes 16,185 images from 196 car models, and Stanford Dogs includes 20,580 images from 120 breeds. \textit{The results on Stanford Cars and Stanford Dogs are in supplementary.} We also extend our evaluation by performing concept learning \cite{koh_concept_2020,huang_concept_2024} experiments on CUB-200-2011 to evaluate whether the learned prototypes can be adapted to produce trustworthy concept-based explanations by following the evaluation protocol introduced in \cite{huang_concept_2024}.

\noindent \textbf{Baselines and Evaluation Metrics.} We establish our main benchmark by comparing against existing ProtoPNets, namely ProtoPNet \cite{chen_this_2019}, Deformable ProtoPNet \cite{donnelly_deformable_2022}, ProtoPool \cite{rymarczyk_interpretable_2022}, TesNet \cite{wang_interpretable_2021} and EvalProtoPNet \cite{huang_evaluation_2023}. We select three self-supervised Vision Transformer backbone as the backbone: DINOv2 ViT-B and ViT-S trained on LVD-142M \cite{oquab_dinov2_2024} and the original DINO trained on ImageNet \cite{caron_emerging_2021}. All the networks are trained with image-level class labels only, and all previous methods were reproduced with official implementations. For evaluation metrics, we report top-1 accuracy for classification. We also quantitatively measure the interpretability of the learned prototypes with the introduced Distinctiveness and Comprehensiveness scores, as well as Consistency and Stability proposed in \cite{huang_evaluation_2023}. Specifically, Consistency evaluates whether each prototype correspond to the same object part across different images, while Stability measures whether each prototype correspond to the same object part when the image is perturbed with noise. We evaluate all methods with $K=3$ and $K=5$. For concept learning, we compare with existing Concept Bottleneck Models \cite{koh_concept_2020,huang_concept_2024}, and evaluate the classification performance as well as the trustworthiness \cite{huang_concept_2024} of the generated concept explanations.

\noindent \textbf{Implementation Details.} In the first stage of training, we set $\beta=0.99$, and $\kappa=0.05$ to perform clustering. When fine-tuning the backbone, we set $\lambda_{\text{PPC}}=0.8$ and the number of added ViT block $m=1$. We trained our framework with 1 epoch for the first stage, and 5 more epochs for the second stage with Adam optimizer, a learning rate of $0.0001$ for the backbone and $1e-6$ for modulation vector $w$. When training previous ProtoPNets, we set $m=3$ as it produces the best results across all previous architectures.

\subsection{Quantitative Results}

We reported the benchmark result of our method against previous ProtoPNets in Table \ref{tab:main-protopnets-results}. Here the Distinctiveness is calculated with $1/4$ of image size. Our method outperformed previous methods in terms of classification accuracy by ($0.16\%$, $2.39\%$) when trained with both the revised DINOv2 and DINO ViT-B backbones, and the accuracy evaluated on ViT-S is comparable to previous works. In addition, while our method yields similar Consistency to \cite{huang_evaluation_2023}, it achieves higher Stability of than existing works in most settings. Nevertheless, the most notable improvement of our framework is the Distinctiveness, which surpasses previous methods by ($4.83\%$, $3.49\%$, $8.93\%$) when setting 5 prototypes for each class. This indicates that our framework is more capable of utilizing the strong part representation encoded in self-supervised Vision Transformers to discover a diverse set of prototypes. Finally, the prototypes learned by our framework can be directly adapted to provide concept-based explanations, as shown by the results in Table \ref{tab:concept-learning-results}. Our method yields comparable trustworthiness score with higher classification performance than \cite{huang_concept_2024}, and Concept Bottleneck Models constructed naively on top of ViT backbones failed to produce trustworthy explanations despite their higher accuracy.

\begin{figure}
  \centering
  \begin{subfigure}[b]{\columnwidth}
    \includegraphics[width=\columnwidth]{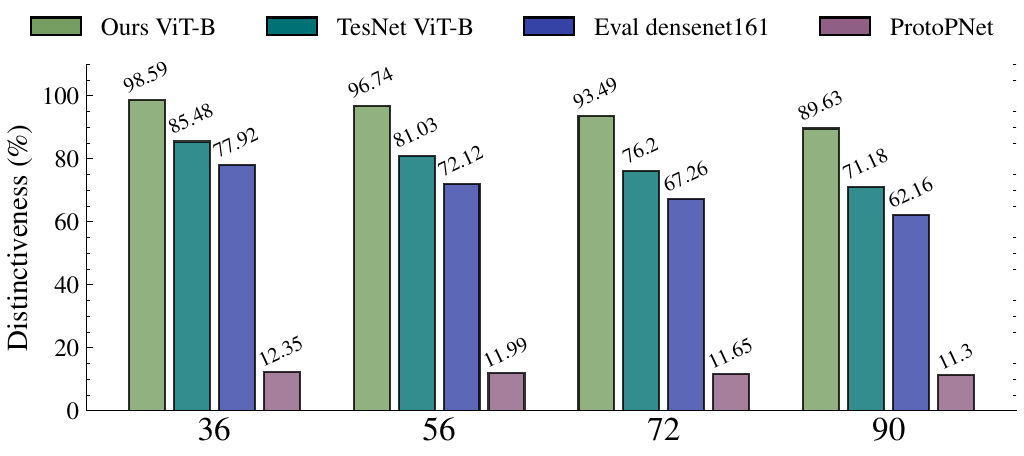}
  \end{subfigure}
  \begin{subfigure}[b]{\columnwidth}
    \includegraphics[width=\columnwidth]{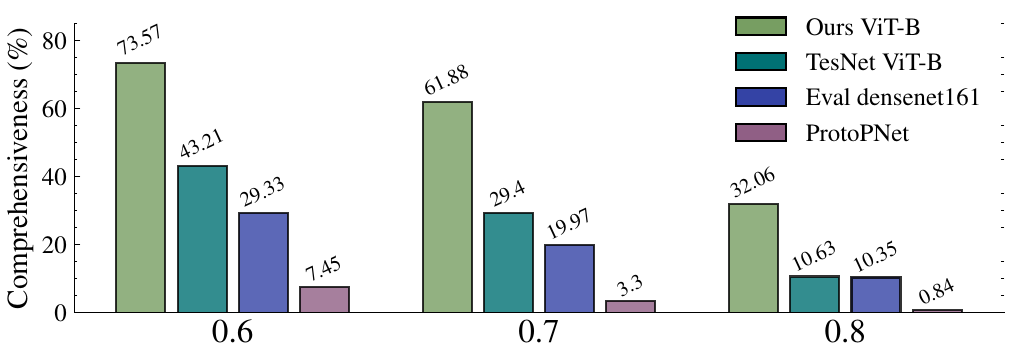}
  \end{subfigure}
  \caption[architecture]{Distinctiveness (\%) and Comprehensiveness (\%) across various ProtoPNets, evaluated on different box sizes and thresholds. All methods besides \cite{huang_evaluation_2023} are trained with DINOv2 ViT-B.}
  \label{fig:distinctiveness-comprehensiveness}
\end{figure}

\begin{table}
  \centering
  \resizebox{0.8\columnwidth}{!}{
    \begin{tabular}{l||cc||cc}
      & \multicolumn{2}{c||}{\textbf{ViT-B}} & \multicolumn{2}{c}{\textbf{ViT-S}} \\
      & Loc. & Acc. & Loc. & Acc. \\
      \hline
      Vanilla CBM & 15.63 & 90.06 & N/A & 88.43 \\
      \hline
      Huang et. al. & 45.95 & 87.43 & 47.33 & 84.79 \\
      Ours (K=3) & 41.18 & 87.04 & 39.88 & 84.55 \\
      Ours (K=5) & 47.04 & 88.18 & 44.97 & 85.47
    \end{tabular}
  }
  \caption{Concept trustworthiness (\%) and classification accuracy (\%) of a CBM constructed based on our framework, in comparison with previous methods on CUB-200-2011 dataset. All methods are trained with DINOv2 ViTs as the backbone.}
  \label{tab:concept-learning-results}
\end{table}


\noindent \textbf{Distinctiveness and Comprehensiveness of Prototypes.} We provide a more detailed breakdown on the diversity of prototypes learned by various ProtoPNets, by selecting a few architectures and evaluate the Distinctiveness with varying bounding box sizes. The results are shown in Figure \ref{fig:distinctiveness-comprehensiveness}. Notably, the prototypes learned by the vanilla ProtoPNet has poor diversity, which means that the network is frequently performing classification by comparing latent image patches with mostly the same object part, which hinders the interpretability. EvalProtoPNet and TesNet both yield much better scores than vanilla ProtoPNet. This phenomenon may be attributed to their orthogonality regularization. For EvalProtoPNet, the Distinctiveness is similar when trained with ViT and CNN backbones. However, our framework still yields significantly better distinctiveness scores across all bounding boxes sizes. This indicates that prototypes of each class generated by our framework attend to a more diverse set of object parts, which result in less overlap between the explanations provided, thereby enhancing the interpretability. In the bottom of Figure \ref{fig:visualization-comparison}, we also show the Comprehensiveness score of architectures evaluated with varying thresholds. The original ProtoPNet fail to generate explanations that cover multiple parts of the object, resulting in very low scores. Among previous works, TesNet yields significantly better foreground coverage, which could also potentially be explained by additional regularizations enforced on the prototypes. Nevertheless, our framework is still able to concepts that provides better coverage of various parts of the objects, achieving significantly higher Comprehensiveness.

\subsection{Qualitative Results}

\begin{figure*}[t]
  \centering
  \includegraphics[width=\textwidth]{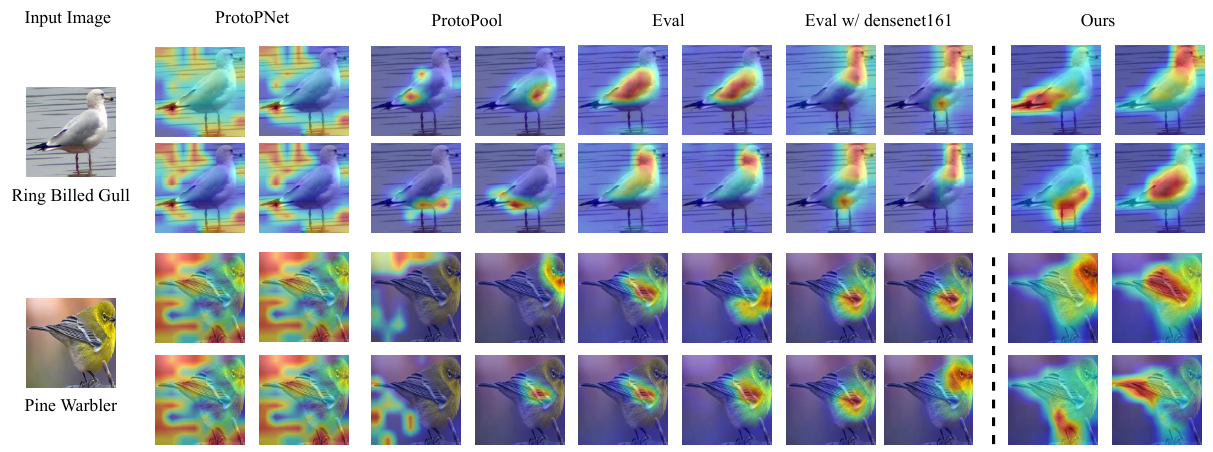}
  \caption[architecture]{Visualization on the same sample across various ProtoPNet architectures. All models are trained using the same DINOv2 ViT-B backbone with $K=5$ unless annotated.}
  \label{fig:visualization-comparison}
\end{figure*}

We presented a comparison of visualization results across ProtoPNet architectures Figure \ref{fig:visualization-comparison}, which displays the activation map of top-$4$ prototypes in heat maps. On the left, it can be seen that the vanilla ProtoPNet has provided poor interpretability for the two examples when trained with ViT backbones. The most activated areas are mostly outside the foreground object and are scattered across the background. For ProtoPool, it has learned more prototypes that attend to various areas within the object, however it still produces some explanations that highlight the background. On the other hand, EvalProtoPNet produces better visual explanations among previous ProtoPNets as most prototypes attend to some different objects parts. However, it lacks diversity as many of them correspond to same object part. For each example, at least two activation map out of four highlight the same concept. This observation is consistent when EvalProtoPNet is trained with convolutional backbones, such as densenet161. In comparison, the prototypes learned by our framework is able to capture different parts of the bird, including head, wing, tail and legs for both examples, resulting in the most conceptually divers set of explanations.

\subsection{Ablation Study}

In this section, we ablate various important design decisions we made in our framework. More experimental results are in the supplementary material. As shown in Table \ref{tab:ablation-modules}, all components of our framework are crucial:

\begin{table}[t]
  \centering
  \resizebox{\columnwidth}{!}{
    \begin{tabular}{l||ccc|c}
      & Con. & Sta. & Dis. & Class. \\
      \hline
      No constraints & 4.09 & 75.50 & 49.47 & 9.25\\
      No foreground extraction & 28.40 & 82.33 & 92.69 & 64.29 \\
      \hline
      No feature space tuning & 63.74 & 83.37 & 96.07 & 77.96 \\
      No loss & 55.61 & 77.67 & 80.41 & 90.78 \\
      Full Model & \textbf{66.4} & \textbf{82.97} & \textbf{96.74} & \textbf{91.02}
    \end{tabular}
  }
  \caption{The effect of design choices made within our framework on the performance and interpretability of our method.}
  \label{tab:ablation-modules}
  \vspace{-1em}
\end{table}

\noindent \textbf{Non-parametric Prototypes.} The first two rows display the importance of \textit{unique-assignment} and \textit{equipartition} constraints when learning non-parametric prototypes. Without these constraints, the clustering algorithm could not properly group semantically similar patch tokens into one cluster and the resulting prototypes are incapable of performing classification. Secondly, without the foreground extraction module, all the background tokens will be involved in the prototype learning stage, which significantly decrease the quality of the prototypes. However, when the full first stage of training is completed, that is, the feature extracted from frozen DINO ViT backbone is clustered with the constraints, the learned prototypes are already capable of recognizing objects and yielding decent interpretability.

\noindent \textbf{Feature Space Fine-tuning.} Our proposed feature space tuning greatly improves the classification accuracy of the model by separating the feature space of different classes while grounding to the part prototypes of each class. However, fine-tuning the added ViT blocks without any guidance reduces the Consistency, Stability as well as Distinctiveness, due to the disturbance of the well-structured feature space of DINO backbone. This makes $\mathcal{L}_{PPC}$ crucial for the fine-tuning stage. By enforcing each latent patch to be more similar to their corresponding part-prototype, it further improves the accuracy while recover model's performance as well as overall interpretability in terms Consistency, Stability as well as Distinctiveness.



\section{Conclusion}
\label{sec:conclusion}

In this work, we presented a framework that provides part-based interpretability for image classification. We identify existing ProtoPNets' limitation of generating visual explanations that are often based on repetitive object parts, which limits their explainability. We mitigate this drawback by adopting non-parametric prototypes to represent diverse and comprehensive object parts. We propose Distinctiveness and Comprehensiveness scores to quantitatively evaluate the quality of explanation provided by various ProtoPNets, complementing the existing suite of metrics, allowing future works to perform evaluation more holistically. We envision this study to serve as a foundation for future work towards stronger part-based interpretable networks.


{
    \small
    \bibliographystyle{ieeenat_fullname}
    \bibliography{main}

\begin{thebibliography}{50}
\providecommand{\natexlab}[1]{#1}
\providecommand{\url}[1]{\texttt{#1}}
\expandafter\ifx\csname urlstyle\endcsname\relax
  \providecommand{\doi}[1]{doi: #1}\else
  \providecommand{\doi}{doi: \begingroup \urlstyle{rm}\Url}\fi

\bibitem[Aamodt and Plaza()]{aamodt_case-based_1994}
Agnar Aamodt and Enric Plaza.
\newblock Case-based reasoning: Foundational issues, methodological variations, and system approaches.
\newblock 7\penalty0 (1):\penalty0 39--59.

\bibitem[Amir et~al.(2021)Amir, Gandelsman, Bagon, and Dekel]{amir_deep_2022}
Shir Amir, Yossi Gandelsman, Shai Bagon, and Tali Dekel.
\newblock Deep vit features as dense visual descriptors.
\newblock \emph{arXiv preprint arXiv:2112.05814}, 2\penalty0 (3):\penalty0 4, 2021.

\bibitem[Awais et~al.(2025)Awais, Naseer, Khan, et~al.]{awais_foundational_2023}
Muhammad Awais, Muzammal Naseer, Salman Khan, et~al.
\newblock Foundation models defining a new era in vision: a survey and outlook.
\newblock \emph{TPAMI}, 2025.

\bibitem[Bafghi et~al.(2024)Bafghi, Harilal, Monteleoni, and Raissi]{bafghi_parameter_2024}
Reza~Akbarian Bafghi, Nidhin Harilal, Claire Monteleoni, and Maziar Raissi.
\newblock Parameter efficient fine-tuning of self-supervised vits without catastrophic forgetting.
\newblock In \emph{CVPR}, pages 3679--3684, 2024.

\bibitem[Carmichael et~al.(2024)Carmichael, Lohit, Cherian, et~al.]{carmichael_pixel-grounded_2024}
Zachariah Carmichael, Suhas Lohit, Anoop Cherian, et~al.
\newblock Pixel-grounded prototypical part networks.
\newblock In \emph{WACV}, pages 4768--4779, 2024.

\bibitem[Caron et~al.(2021)Caron, Touvron, Misra, et~al.]{caron_emerging_2021}
Mathilde Caron, Hugo Touvron, Ishan Misra, et~al.
\newblock Emerging properties in self-supervised vision transformers.
\newblock In \emph{ICCV}, pages 9650--9660, 2021.

\bibitem[Chen et~al.(2019)Chen, Li, Tao, Barnett, Rudin, and Su]{chen_this_2019}
Chaofan Chen, Oscar Li, Daniel Tao, Alina Barnett, Cynthia Rudin, and Jonathan~K Su.
\newblock This {Looks} {Like} {That}: {Deep} {Learning} for {Interpretable} {Image} {Recognition}.
\newblock In \emph{NeurIPS}, 2019.

\bibitem[Chen et~al.()Chen, Kornblith, Norouzi, and Hinton]{chen_simple_2020}
Ting Chen, Simon Kornblith, Mohammad Norouzi, and Geoffrey Hinton.
\newblock A simple framework for contrastive learning of visual representations.
\newblock In \emph{ICML}, pages 1597--1607.

\bibitem[Cuturi()]{cuturi_sinkhorn_2013}
Marco Cuturi.
\newblock Sinkhorn distances: Lightspeed computation of optimal transport.
\newblock In \emph{NeurIPS}. Curran Associates, Inc.

\bibitem[Dataset(2011)]{dataset2011novel}
E Dataset.
\newblock Novel datasets for fine-grained image categorization.
\newblock In \emph{CVPR Workshop}, page~2. Citeseer, 2011.

\bibitem[Donnelly et~al.(2022)Donnelly, Barnett, and Chen]{donnelly_deformable_2022}
Jon Donnelly, Alina~Jade Barnett, and Chaofan Chen.
\newblock Deformable {ProtoPNet}: An interpretable image classifier using deformable prototypes.
\newblock In \emph{CVPR}, pages 10265--10275, 2022.

\bibitem[Espinosa~Zarlenga et~al.(2022)Espinosa~Zarlenga, Barbiero, et~al.]{espinosa_zarlenga_concept_2022}
Mateo Espinosa~Zarlenga, Pietro Barbiero, et~al.
\newblock Concept embedding models: Beyond the accuracy-explainability trade-off.
\newblock 35:\penalty0 21400--21413, 2022.

\bibitem[Fan et~al.(2022{\natexlab{a}})Fan, Ding, Fan, et~al.]{fan2022grainspace}
Lei Fan, Yiwen Ding, Dongdong Fan, et~al.
\newblock Grainspace: A large-scale dataset for fine-grained and domain-adaptive recognition of cereal grains.
\newblock In \emph{CVPR}, pages 21116--21125, 2022{\natexlab{a}}.

\bibitem[Fan et~al.(2022{\natexlab{b}})Fan, Sowmya, Meijering, and Song]{fan2022cancer}
Lei Fan, Arcot Sowmya, Erik Meijering, and Yang Song.
\newblock Cancer survival prediction from whole slide images with self-supervised learning and slide consistency.
\newblock \emph{IEEE TMI}, 42\penalty0 (5):\penalty0 1401--1412, 2022{\natexlab{b}}.

\bibitem[Fan et~al.(2023{\natexlab{a}})Fan, Ding, Fan, et~al.]{fan2023identifying}
Lei Fan, Yiwen Ding, Dongdong Fan, et~al.
\newblock Identifying the defective: Detecting damaged grains for cereal appearance inspection.
\newblock In \emph{ECAI}, pages 660--667. IOS Press, 2023{\natexlab{a}}.

\bibitem[Fan et~al.(2023{\natexlab{b}})Fan, Fan, Ding, et~al.]{fan2023av4gainsp}
Lei Fan, Dongdong Fan, Yiwen Ding, et~al.
\newblock Av4gainsp: An efficient dual-camera system for identifying defective kernels of cereal grains.
\newblock \emph{IEEE Robotics and Automation Letters}, 9\penalty0 (1):\penalty0 851--858, 2023{\natexlab{b}}.

\bibitem[Fan et~al.(2025)Fan, Fan, Ding, et~al.]{fan2025grainbrain}
Lei Fan, Dongdong Fan, Yiwen Ding, et~al.
\newblock Grainbrain: Multiview identification and stratification of defective grain kernels.
\newblock \emph{IEEE TII}, 2025.

\bibitem[He et~al.()He, Fan, Wu, Xie, and Girshick]{he_momentum_2020}
Kaiming He, Haoqi Fan, Yuxin Wu, Saining Xie, and Ross Girshick.
\newblock Momentum contrast for unsupervised visual representation learning.
\newblock In \emph{CVPR}, pages 9729--9738.

\bibitem[Huang et~al.(2023)Huang, Xue, Huang, Zhang, Song, Jing, and Song]{huang_evaluation_2023}
Qihan Huang, Mengqi Xue, Wenqi Huang, Haofei Zhang, Jie Song, Yongcheng Jing, and Mingli Song.
\newblock Evaluation and improvement of interpretability for self-explainable part-prototype networks.
\newblock In \emph{ICCV}, pages 2011--2020, 2023.

\bibitem[Huang et~al.(2024)Huang, Song, Hu, et~al.]{huang_concept_2024}
Qihan Huang, Jie Song, Jingwen Hu, et~al.
\newblock On the {Concept} {Trustworthiness} in {Concept} {Bottleneck} {Models}.
\newblock \emph{AAAI}, 38\penalty0 (19):\penalty0 21161--21168, 2024.

\bibitem[Huang and Li()]{huang_interpretable_2020}
Zixuan Huang and Yin Li.
\newblock Interpretable and accurate fine-grained recognition via region grouping.
\newblock In \emph{CVPR}, pages 8662--8672.

\bibitem[Kirillov et~al.()Kirillov, Mintun, Ravi, et~al.]{kirillov_segment_2023}
Alexander Kirillov, Eric Mintun, Nikhila Ravi, et~al.
\newblock Segment anything.
\newblock In \emph{ICCV}, pages 4015--4026.

\bibitem[Koh et~al.(2020)Koh, Nguyen, Tang, Mussmann, Pierson, Kim, and Liang]{koh_concept_2020}
Pang~Wei Koh, Thao Nguyen, Yew~Siang Tang, Stephen Mussmann, Emma Pierson, Been Kim, and Percy Liang.
\newblock Concept bottleneck models.
\newblock In \emph{ICML}, pages 5338--5348. {PMLR}, 2020.

\bibitem[Krause et~al.(2013)Krause, Stark, Deng, and Fei-Fei]{krause_3d_2013}
Jonathan Krause, Michael Stark, Jia Deng, and Li Fei-Fei.
\newblock 3d object representations for fine-grained categorization.
\newblock In \emph{ICCV workshops}, pages 554--561, 2013.

\bibitem[Lu et~al.(2024)Lu, Liu, Wang, Han, Cui, Cao, Zhang, Chen, and Fan]{lu_promotion_2024}
Yawen Lu, Dongfang Liu, Qifan Wang, Cheng Han, Yiming Cui, Zhiwen Cao, Xueling Zhang, Yingjie~Victor Chen, and Heng Fan.
\newblock {ProMotion}: Prototypes as motion learners.
\newblock In \emph{CVPR}, pages 28109--28119, 2024.

\bibitem[Oquab et~al.(2024)Oquab, Darcet, Moutakanni, et~al.]{oquab_dinov2_2024}
Maxime Oquab, Timothée Darcet, Théo Moutakanni, et~al.
\newblock {DINOv2}: {Learning} {Robust} {Visual} {Features} without {Supervision}.
\newblock \emph{TMLR}, 2024.

\bibitem[Park et~al.()Park, Kim, Heo, Kim, and Yun]{park_what_2022}
Namuk Park, Wonjae Kim, Byeongho Heo, Taekyung Kim, and Sangdoo Yun.
\newblock What do self-supervised vision transformers learn?
\newblock In \emph{ICLR}.

\bibitem[Qin et~al.(2023)Qin, Han, Wang, Nie, Yin, and Xiankai]{qin_unified_2023}
Zheyun Qin, Cheng Han, Qifan Wang, Xiushan Nie, Yilong Yin, and Lu Xiankai.
\newblock Unified 3d segmenter as prototypical classifiers.
\newblock \emph{NeurIPS}, 36:\penalty0 46419--46432, 2023.

\bibitem[Radford et~al.()Radford, Kim, Hallacy, et~al.]{radford_learning_2021}
Alec Radford, Jong~Wook Kim, Chris Hallacy, et~al.
\newblock Learning transferable visual models from natural language supervision.
\newblock In \emph{ICML}, pages 8748--8763. {PMLR}.

\bibitem[Rymarczyk et~al.(2022)Rymarczyk, Struski, Górszczak, et~al.]{rymarczyk_interpretable_2022}
Dawid Rymarczyk, Łukasz Struski, Michał Górszczak, et~al.
\newblock Interpretable image classification with differentiable prototypes assignment.
\newblock In \emph{ECCV}, pages 351--368, 2022.

\bibitem[Saha and Maji()]{saha_particle_2023}
Oindrila Saha and Subhransu Maji.
\newblock {PARTICLE}: Part discovery and contrastive learning for fine-grained recognition.
\newblock In \emph{ICCV}, pages 167--176.

\bibitem[Selvaraju et~al.()Selvaraju, Cogswell, Das, Vedantam, Parikh, and Batra]{selvaraju_grad-cam_2017}
Ramprasaath~R. Selvaraju, Michael Cogswell, Abhishek Das, Ramakrishna Vedantam, Devi Parikh, and Dhruv Batra.
\newblock Grad-{CAM}: Visual explanations from deep networks via gradient-based localization.
\newblock In \emph{ICCV}, pages 618--626.

\bibitem[Sim{\'e}oni et~al.(2021)Sim{\'e}oni, Puy, Vo, Roburin, Gidaris, Bursuc, P{\'e}rez, Marlet, and Ponce]{simeoni_localizing_2021}
Oriane Sim{\'e}oni, Gilles Puy, Huy~V Vo, Simon Roburin, Spyros Gidaris, Andrei Bursuc, Patrick P{\'e}rez, Renaud Marlet, and Jean Ponce.
\newblock Localizing objects with self-supervised transformers and no labels.
\newblock In \emph{BMVC}, 2021.

\bibitem[Sun et~al.(2024)Sun, Wang, Tan, Dong, Ma, Li, and Gong]{sun2024eggen}
Zhenhong Sun, Junyan Wang, Zhiyu Tan, Daoyi Dong, Hailan Ma, Hao Li, and Dong Gong.
\newblock {EGGen: Image Generation with Multi-entity Prior Learning through Entity Guidance}.
\newblock In \emph{ACM MM}, pages 6637--6645, 2024.

\bibitem[Tan et~al.(2024)Tan, Zhou, and Chen]{tan_post-hoc_2024}
Andong Tan, Fengtao Zhou, and Hao Chen.
\newblock Post-hoc part-prototype networks.
\newblock In \emph{ICML}, 2024.

\bibitem[van~der Klis et~al.()van~der Klis, Alaniz, Mancini, Dantas, Ienco, Akata, and Marcos]{van_der_klis_pdisconet_2023}
Robert van~der Klis, Stephan Alaniz, Massimiliano Mancini, Cassio~F. Dantas, Dino Ienco, Zeynep Akata, and Diego Marcos.
\newblock {PDiscoNet}: Semantically consistent part discovery for fine-grained recognition.
\newblock In \emph{ICCV}, pages 1866--1876.

\bibitem[Wah et~al.(2011)Wah, Branson, Welinder, Perona, and Belongie]{WahCUB_200_2011}
C. Wah, S. Branson, P. Welinder, P. Perona, and S. Belongie.
\newblock Technical Report CNS-TR-2011-001, California Institute of Technology, 2011.

\bibitem[Wang et~al.(2023)Wang, Li, Nakashima, and Nagahara]{wang_learning_2023}
Bowen Wang, Liangzhi Li, Yuta Nakashima, and Hajime Nagahara.
\newblock Learning bottleneck concepts in image classification.
\newblock In \emph{CVPR}, pages 10962--10971, 2023.

\bibitem[Wang et~al.(2021)Wang, Liu, Wang, and Jing]{wang_interpretable_2021}
Jiaqi Wang, Huafeng Liu, Xinyue Wang, and Liping Jing.
\newblock Interpretable image recognition by constructing transparent embedding space.
\newblock In \emph{ICCV}, pages 895--904, 2021.

\bibitem[Wang et~al.(2024)Wang, Sun, Tan, Chen, Chen, Li, Zhang, and Song]{wang2024towards}
Junyan Wang, Zhenhong Sun, Zhiyu Tan, Xuanbai Chen, Weihua Chen, Hao Li, Cheng Zhang, and Yang Song.
\newblock Towards effective usage of human-centric priors in diffusion models for text-based human image generation.
\newblock In \emph{CVPR}, pages 8446--8455, 2024.

\bibitem[Wang et~al.()Wang, Han, Zhou, and Liu]{wang_visual_2023}
Wenguan Wang, Cheng Han, Tianfei Zhou, and Dongfang Liu.
\newblock Visual recognition with deep nearest centroids.
\newblock In \emph{ICLR}.

\bibitem[Wu et~al.()Wu, Xiong, Yu, and Lin]{wu_unsupervised_2018}
Zhirong Wu, Yuanjun Xiong, Stella~X. Yu, and Dahua Lin.
\newblock Unsupervised feature learning via non-parametric instance discrimination.
\newblock In \emph{CVPR}, pages 3733--3742.

\bibitem[Xie et~al.({\natexlab{a}})Xie, Geng, Hu, Zhang, Hu, and Cao]{xie_revealing_2023}
Zhenda Xie, Zigang Geng, Jingcheng Hu, Zheng Zhang, Han Hu, and Yue Cao.
\newblock Revealing the dark secrets of masked image modeling.
\newblock In \emph{CVPR}, pages 14475--14485, {\natexlab{a}}.

\bibitem[Xie et~al.({\natexlab{b}})Xie, Zhang, Cao, Lin, Bao, Yao, Dai, and Hu]{xie_simmim_2022}
Zhenda Xie, Zheng Zhang, Yue Cao, Yutong Lin, Jianmin Bao, Zhuliang Yao, Qi Dai, and Han Hu.
\newblock {SimMIM}: A simple framework for masked image modeling.
\newblock In \emph{CVPR}, pages 9653--9663, {\natexlab{b}}.

\bibitem[Xue et~al.(2024)Xue, Huang, Zhang, Hu, Song, Song, and Jin]{xue_protopformer_2022}
Mengqi Xue, Qihan Huang, Haofei Zhang, Jingwen Hu, Jie Song, Mingli Song, and Canghong Jin.
\newblock Protopformer: concentrating on prototypical parts in vision transformers for interpretable image recognition.
\newblock In \emph{IJCAI}, pages 1516--1524, 2024.

\bibitem[Yang et~al.()Yang, Liu, Li, Jiao, and Ye]{yang_prototype_2020}
Boyu Yang, Chang Liu, Bohao Li, Jianbin Jiao, and Qixiang Ye.
\newblock Prototype mixture models for few-shot semantic segmentation.
\newblock In \emph{ECCV}, pages 763--778. Springer International Publishing.

\bibitem[Zhang et~al.(2024)Zhang, Du, Yan, and Zhang]{zhang2024decoupling}
Rui Zhang, Xingbo Du, Junchi Yan, and Shihua Zhang.
\newblock The decoupling concept bottleneck model.
\newblock \emph{TPAMI}, 2024.

\bibitem[Zhou et~al.({\natexlab{a}})Zhou, Khosla, Lapedriza, Oliva, and Torralba]{zhou_learning_2016}
Bolei Zhou, Aditya Khosla, Agata Lapedriza, Aude Oliva, and Antonio Torralba.
\newblock Learning deep features for discriminative localization.
\newblock In \emph{CVPR}, pages 2921--2929, {\natexlab{a}}.

\bibitem[Zhou and Wang(2024)]{zhou_prototype-based_2024}
Tianfei Zhou and Wenguan Wang.
\newblock Prototype-based semantic segmentation.
\newblock \emph{TPAMI}, 2024.

\bibitem[Zhou et~al.({\natexlab{b}})Zhou, Wang, Konukoglu, and Van~Gool]{zhou_rethinking_2022}
Tianfei Zhou, Wenguan Wang, Ender Konukoglu, and Luc Van~Gool.
\newblock Rethinking semantic segmentation: A prototype view.
\newblock In \emph{CVPR}, pages 2582--2593, {\natexlab{b}}.

\end{thebibliography}
}

\end{document}